# Describing Upper Body Motions based on the Labanotation for Learning-from-Observation Robots

Katsushi Ikeuchi, Zengqiang Yan, Zhaoyuan Ma, Yoshihiro Sato, Minako Nakamura, Shunsuke Kudoh

*Abstract*— **We have been developing a paradigm, which we refer to as Learning-from-observation, for a robot to automatically acquire what-to-do through observation of human performance. Since a simple mimicking method to repeat exact joint angles does not work due to the kinematic and dynamic difference between a human and a robot, the method introduces an intermediate symbolic representation, task models, to conceptually represent what-to-do through observation. Then, these task models are mapped appropriate robot motions depending on each robot hardware. This paper presents task models, designed based on the Labanotation, for upper body movements of humanoid robots. Given a human motion sequence, we first analyze the motions of the upper body, and extract certain fixed poses at certain key frames. These key poses are translated into states represented by Labanotation symbols. Then, task models, identified from the state transitions, are mapped to robot movements on a particular robot hardware. Since the task models based on Labanotation are independent from different robot hardware, we can share the same observation module; we only need task mapping modules depending on different robot hardware. The system was implemented and demonstrated that three different robots can automatically mimic human upper body motions with satisfactory level of resemblance.**

## I. INTRODUCTION

RECENTLY, robot application areas have been drastically increasing. Traditionally, their applications were rather limited in industrial applications. Recently, robots have been used in other areas including family service [1], medical applications [2,3], and even defense applications [4,5]. Along

K. Ikeuchi and Z. Ma are with Microsoft Research Asia, Microsoft, Beijing, China (e-mail: katsuike@microsoft.com; v-zhma@micorsoft.com )

Z. Yan is with Huazhong University of Science and Technology, Wuhan, China (e-mail: ZengqiangYan@hust.edu.cn).

Y. Sato is with the University of Tokyo, Tokyo, Japan (e-mail: yoshi@cvl.iis.u-tokyo.ac.jp)

M. Nakamura is with Ochanomizu University, Tokyo, Japan (e-mail: nakamura.minako@ocha.ac.jp).

S. Kudoh is with the University of Electro-Communications, Tokyo, Japan (e-mail: kudoh@is.uec.ac.jp).

this line of increasing trend, one of the imminent issues is how to program such robots in efficient manners.

We have been working on the learning-from-observation paradigm to overcome the burden of programing efforts [6, 7]. If we can make a robot to be able to learn how to perform a task just from observing human performance of the same task, we can drastically decrease the cost of programming. Toward this goal, we proposed the task-and-skill model framework so as to separate a common component, what to do, tasks from personal variegations, skills how to do [6,7].

Under this task-skill modeling, we have explored necessary and sufficient sets of states in various domains, including two cubes [6], two polyhedral [7], mechanical parts [8] and knotting rope world [9]. In 2007, we demonstrated this task-skill model for a humanoid robot to dance a Japanese folk dance, Aizubanda-san dance [10]. Although the robot can successful dance a Japanese folk dance so as to attract a large audience, we can only define task models for the lower body. The definition of upper body task models has been an open issue since then.

In the robotics field, many researchers have developed methods to adapt human motion to a humanoid robot. Riley produced a dancing motion of a humanoid robot by converting human motion data, by a motion capture system, into joint trajectories of the robot [11]. For the same purpose, Pollard proposed a method for constraining given joint trajectories within mechanical limitations of the joints [12]. For biped humanoid robots, Tamiya proposed a method that enables a robot to follow given motion trajectories while keeping body balance [13]. Kagami extended the method so that it allows the changes of supporting legs [14]. Yamane proposed a dynamics filter, which converts a physically inconsistent motion into a consistent one for a given body [15]. These works are mainly concern with how to create a new trajectory of a joint within a given physical constraint; there is no attempt to describe global motion structures in symbolic representations.

With regard to dance performance, Kuroki enabled an actual biped humanoid to stably perform dance motions that include dynamic-style steps [16]. Nakaoka also developed a similar dancing robot based on the software Choreonoid [17]. These robots are manually coded and no analysis exists. Kawato's group proposes a humanoid robot to learn Okinawa-teodori based on neural network approach [18]. The result is interesting, however due to the bottom-up nature of the learning mechanism, it is difficult to conduct the analysis of dance structure for further preservation purpose. Kosuge proposes a dance-partner robot for western dance. The robot performs excellent dance based on the partner's motion [19]. Okuno's group developed a humanoid robot to step along with the music beat [20]. The motion is limited on stepping actions.

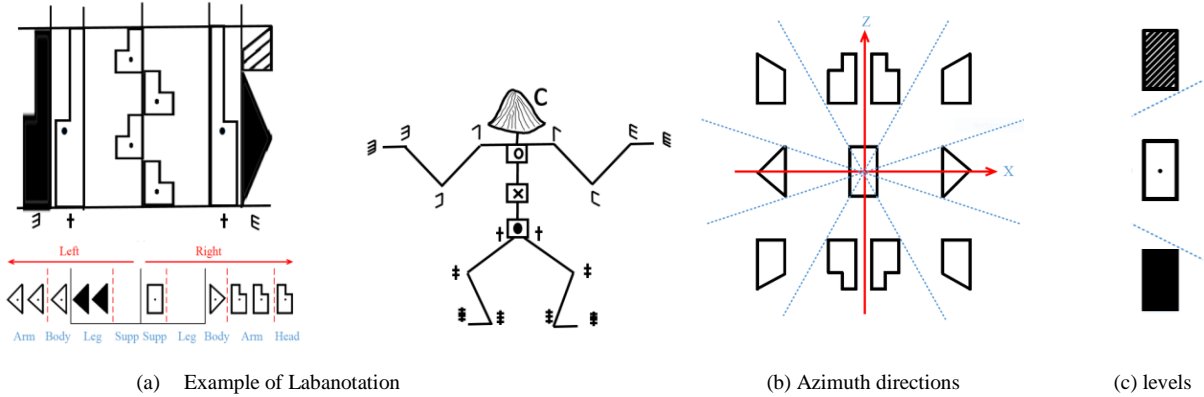

(a) Example of Labanotation          (b) Azimuth directions      (c) levels

Fig. 1. Labanotation. In a Labanotation score, the time passes from bottom to top. Each column in the Labanotation score corresponds to one part of a human body as indicated in the left drawing. Each symbol denotes the direction in its shape and the level in painted pattern inside of the symbol.

In contract to these earlier attempts, this paper proposes Labanotation to describe upper body task models for a humanoid robot. The Labanotation has been used in dance community to record human dances [21,22]. Some robotics researchers also proposed to use this Labanotation as the basis of the robot language design [23]. We will use this Labanotation for describing states of upper body motions and design task models for any upper body motions of the learning-from-observation humanoid robot.

The observation module of our system is related with human action recognition. Recently research on recognize human actions from visual observation have been developed and well-studied. Representative work includes: actionlet ensemble model [24], convolutional neural network [25], [26], trajectories [27], [28], and motion characteristics [29], [30]. There are a couple of databases to evaluate performance of recognition systems, including UCF sports datasets [31], Stanford Olympics datasets [32], and Hollywood movie data sets [33]. However, those methods are mainly concerns on categorization of human actions such as biking, climbing stairs, jumping roping etc. In fact, there is no notion of necessary and sufficient issues in those database and recognition. It is unclear for what purpose such categorization is necessary beyond necessity of video-surveillance tasks. It is also true that such recognition results, unfortunately, cannot be used for creating robot actions, either. The description is the necessary condition, but, it is not the sufficient condition.

Bobick defines human action recognition into three categories: "movement," "activity," and "action" recognition. Among these three categories, movement recognition is closely related with our task recognition [34]. Bobick defines that a movement is "a motion whose execute is consistent and easily characterized by a definite space." We would like to re-define a movement as "a motion with a clear purpose to generate one state transition in one particular action domain." Our task models are defined to specify corresponding movements to create state transitions as their purposes.

One of the imminent issues is, then, how to define states. These states are characterized in various domains. In fact, we have been exploring this necessary and sufficient sets of states in various human action domains, including polyhedral world [7] and lower-body dance motions [10]. This paper designs task models for upper body motions based on the Labanotation.

The following is the organization of this paper. Chapter 2 explains the central concept, Labanotation, and how to convert Kinect output into Labanotation. Chapter 3 is a mapping routine to map Labanotation into robot actions, and demonstrate such system on multiple robot hardware. Chapter 4 concludes this paper.

## II. Labanotation and its extraction

### A. Labanotation

Labanotation is developed by Rudlf V. Laban in early 20th century [21]. Labanotation scores resemble to music scores. Fig.1 (a) shows an example of a Labanotation score. In a music score, the time passes along the horizontal direction from left to right. In a Labanotation score, the time passes along the vertical direction from bottom to top. In a music score, each row line corresponds to a certain frequency, a music scale, and a symbol corresponds whether such sound appears or not at that moment. In a Labanotation score, each column corresponds one body part and a symbol represents to which direction that body part faces at that time.

A Labanotation score is the necessary and sufficient condition to describe one piece of dance as in the same sense that a music score is the necessary and sufficient condition to describe a piece of music. Any musician ends up to record a common music score from listening the same music piece. Any musician ends up to play (or reconstruct) such a common music piece based on the same music score. In the similar way, any expert ends up to record a dance piece into one common Labanotation score. Any dancer ends up to perform (or reconstruct) the same dance piece based on the recorded same Labanotation score. Thus, a Labanotation score is necessary and sufficient condition and one-to-one mapping to a piece of dance.

Each symbol in Labanotation represents the direction of each body part. As shown in Fig 1(b), eleven symbols correspond to eight azimuths and one neutral directions straight up or down, where two symbols are used to specify the same forward/backward direction depending on left or right arm/foot. Some theory says human perception allows roughly seven-plus -minus-three categorizations. The number of the main chords in music is exactly seven. The number of color in rainbow is seven. Along that line of thought, the eight directional digitization may be reasonable to human perception and probably, due to this fact, the dance community has been using this notation for more than a century. Of course, in Labanotation, it is possible to specify the finer directions, if necessary, but it is rare to use such fine grain notations.

The level has been classified into three categories: high, middle, and low. Including top and bottom, the number of the level grain is five. The level is notated as the color inside of the symbol as shown in Fig 1 (c).

For example, the score in Fig.1 (a) can be understand as follows. The central two column represents the support the upper body. During the score period, the left step, the right step, the left step, and the right step occur. During that four steps, right hand stretch out in the middle level, while the left arm stretch out high and low. In this example, we omit other body parts, but, can describe them in the similar way.

### B. Labanotation, Task models and States

In the learning-from-observation paradigm, a task is defined as one robot movement to generate one specific state transition [7]. Here, we define our task recognition as an extension of object recognition. In object recognition, we prepare abstract object models in computer in an off-line mode. In an on-line mode, the computer associates model features with real features, identifies the corresponding abstract objects, and creates such a world representation with instantiate object models. In the similar way, in task recognition, we prepare abstract task models in computer, which associates state transitions with a movement necessary to create such transition in off-line. In on-line, the system detects state transitions, and identifies an abstract task model to associate the state transition observed with a necessary motion to create such transition.

In order to define such task models, we have to define a set of states. For simplicity, let's consider assembly operations of a pair of cubes [7] as shown in Fig.2. In this domain, these two cubes, say A and B, are defined to have four states; "A on top of A," "B on top of A," "A left to B," and "B left to A" as shown in Fig. 2. An assembly operation is characterized such as one to create a transition of contact states among two cubes. For example, one transition is "A left to B" state to "A on top of B" state. To each state transition, we can assign one necessary motion to create such state transition. In this case, "Put-A-on-top-of B" is a necessary action. This association between a state

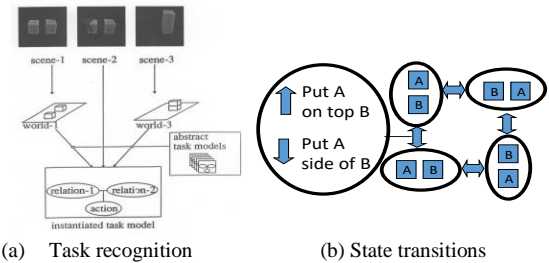

| (a) Task recognition | (b) State transitions |

Fig. 2 Task recognition and States [6]. Abstract task model associates one state transition with a necessary action to create such transition.

transitions with the necessary motion is a task model.

On line, object recognition system identifies current state and previous state. Then, the task recognition system recognizes a state transition, associates a necessary action, and executes those action. This is the concept of task recognition.

The purpose of this task recognition has two folds. By dividing continuous observation space into a discrete set of states and thus task models, we can remove some of observation errors. The second purpose is to separate observation from execution modules so as to be able to share the same observation module, while to have different kinds of mapping routines depending on each robot with different configurations; we can apply the same set of task models to be able to execute by different robot hardware

We will define a Labanotation symbol to represent one state in human motion. This is slightly different from the interpretation of original Labanotation symbols. In the original Labanotation, each symbol is explained as a movement of a body part to reach one particular pose. This paper defines Labanotation symbols to represent the final poses as the result of movements.

We, further, define that one task includes that ending state; however, it does not include the starting state. In the previous example, one task is to move one cube to achieve "A on top of B" state. The task period is defined not to include the start state.

We can define that each body part executes one task in parallel manner. For example, corresponding to one Labanotation symbol such "a black triangle" in the arm column in a Labanotation score in Fig 1(c), the dance performs to stretch the right arm toward the horizontal direction. During that period, the dancer performs four steps. Namely, each body parts, in this example, lower body and arms, executes its own task in parallel manner.

### C. Key frame detection and key pose extraction

Another important component is to decide when one particular task ends. During a sequence of movements, we have to choose one particular pose to be recorder as the end state using a Labanotation symbol. Motion segmentation is necessary to extract such states.

One simple idea is to convert all the poses by human performer at each sampling timings into Labanotation symbols regardless to whether it is same or not, and then to extract any transitions in the Labanotation. We implemented this idea, but, apparently, the resulting score is different from the one given by a Labanotation expert. It is also true that beginners and experts generate different Labanotation scores. Experts converge the same set of Labanotation; the Labanotation society issues certificates to the experts, who is qualified as Labanotation recorder so as for him/her to determine the important styles and to record the Labanotation. Thus, when to record poses into Labanotation scores is another key to be the necessary and sufficient condition of a Labanotation score.

From the discussion with Labanotation experts, brief stops in body movements provide key moments to record such poses. In fact, previously, Shiratori followed this idea, and considered a motion energy function of all the components of human body, i.e. combining motion energy values of all the motions of hands, foot and the head, and determined key frames as local minima of the energy function [21].

As mentioned previously, each body part performs its own task in parallel manner. For example, let's revisit the Labanotation score in Fig1 (a). Apparently, each body components, in this particular example, the foot and the hands are independently represented. While foot will have four steps, left hand only stretch out once. From this, it is apparent that each body part should have its own energy function for motion segmentation; it is not a good idea to sum all the energy values given from the whole body.

For each body part, we design energy functions in the following form.

$$E = f(x, y, z) = E_a(x, y, z) - E_s(x, y, z)$$

where $E_a$ represents the motion acceleration calculated by

$$E_a(x, y, z) = \frac{1}{\sqrt{3}}\sqrt{\left(\frac{\partial^2 x}{\partial^2 t}\right)^2 + \left(\frac{\partial^2 y}{\partial^2 t}\right)^2 + \left(\frac{\partial^2 z}{\partial^2 t}\right)^2},$$

And $E_s$ is the motion speed by

$$E_s(x, y, z) = \frac{1}{\sqrt{3}}\sqrt{\left(\frac{\partial x}{\partial t}\right)^2 + \left(\frac{\partial y}{\partial t}\right)^2 + \left(\frac{\partial z}{\partial t}\right)^2}$$

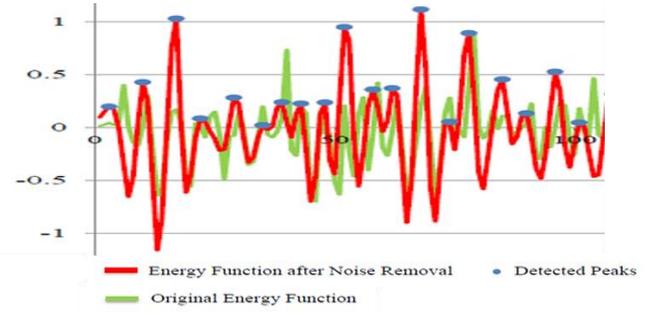

Fig. 3. Results of peak detection in the proposed energy function

and both derivative values are normalized into (0,1). Here, (x,y,z) is the hand position in t.

Energy function of a sequence of actions is provided in Fig.3. Considering that there might exist wrong values and motion blur in the calculation of energy function, a signal smoothing process is accomplished by applying the discrete convolution of a Gaussian based filter to variances x, y and z separately according to

$$f'(x) = f(x) * G(x),$$

with

$$G(x) = \frac{1}{\sigma\sqrt{2\pi}}e^{-\frac{(x-\mu)^2}{2\sigma^2}}$$

Then, energy function after the signal smoothing processing also is shown in Fig.3. In the refined energy function, differences between different points are amplified, which makes it easier to identify the energy of each point.

A key frame is defined as a moment when any of the body parts holds a local minimum energy value. The posture of any body part corresponding to the peaks in the energy function is selected and encoded into a Labanotation symbol. When a multiple parts have local minimum energy values neighboring periods, the average period is used as the key frame.

Visual results of our key frames extraction method are shown in Fig.4. Observing the motion sequence, we can find that selected key frames contain representative gestures at that moments, which we refer to as key poses. From visual observations of key poses, it is more likely to recover the original motion sequence.

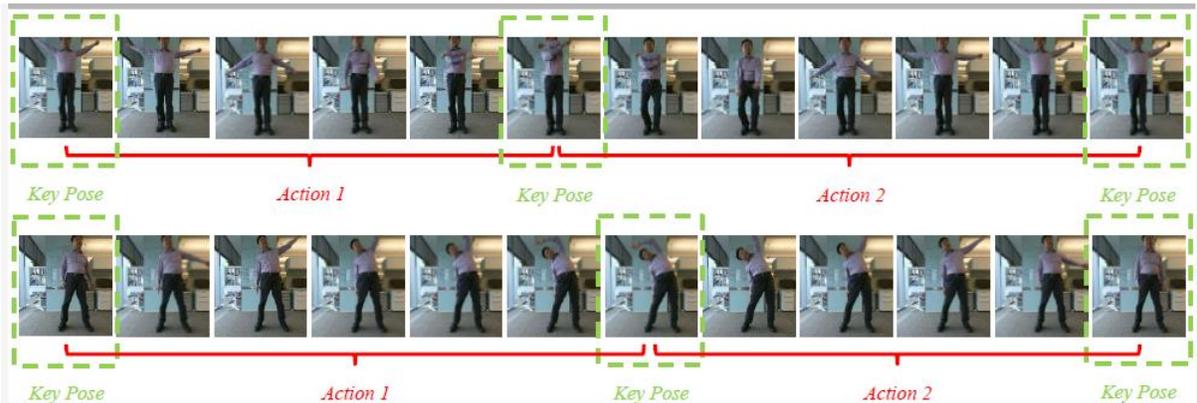

Fig. 4 Extracted key frames and corresponding key poses

### D. Labanotation Encoder

We assume to have a sequence of skeleton output from a Kinect sensor. At each key pose, directions of the human body parts are sampled. As mentioned previously, this encoding is to digitize continuous directional variances into a finite number of directions given by the Labanotation.

We must define the body coordinate system for constructing a Labanotation score. In our system, body coordinate systems of both human and robot are aligned as shown in Figs. 5. To compute a Labanotation symbol of each body part, we first calculate the relative position between the body part and its parent part, and then corresponding Labanotation symbol is selected based on the relative position. Here, for each body part, its parent part is the one near to the origin, namely elbow is the parent of wrist. By calculating angles between the part and the base coordinate system, one Labanotation symbol is selected according to Fig.1 (b). Once Labanotation symbols of body parts are calculated, the key pose can be represented by a combination of Labanotation symbols as an example shown in Fig.1 (a).

Based on the Labanotation, continuous motions can be effectively compressed and encoded. This compression is essential in particular on considering a cloud robot, which is connected to a cloud computer and some motion commands are transmitted through a narrow channel between a robot and a cloud computer. In the Labanotation, the whole motion space is divided into specific symbols, so that any gesture can be classified into a combination of symbols in the reasonable degree of coarseness corresponding to human perception.

### III. MAPPING LABANOTATION TO ROBOT MOVEMENTS

#### A. Labanotation Decoder

A Labanotation decoder maps a Labanotation score to a sequence of motions on a robot. Each robot has different configurations; we prepare Labanotation decoders corresponding to each specific robots. In this section, for the sake of clarity, we will first explain a simple 7 DOF robot as a test bed. We also assume that a Labanotation score has only arm representation. Then, later, we will explain how to extend the method to other complicated cases.

Fig. 5 shows one simple robot with 7 DOFs. As for the motion of the arms, this robot has two DOFs around the shoulder. This robot also has one DOF around the wrist. The head has also two DOFs. The total DOFs of this robot is 7 DOFs. Each joint is controlled by Futaba motor through RD303MR.

The Labanotation decoder is rather simple. We digitize the DOF space in the 8 direction and 3 levels corresponding the Labanotation. The roll space of the shoulder joint is represented as a set of four configurations corresponding to the Labanotation symbols. Due to the limitation of the robot, only frontal gestures is implemented on this robot. Of course, some dance may have more complicated gestures such as move one arm to back ward direction; we ignore such gestures on this simple implementation. All the outside gestures are represented as the boundary gesture. As for pitch direction, following the Labanotation, we digitize the direction into three levels: high, middle, and low. All the possible configurations of the right arm are represented in Fig.6 Then, we can assign Labanotation symbols to those configurations. In this simple example, since the DOFs in Labanotation is same as the DOFs of the robot, simple mapping of configuration of body parts work.

In a general case, we have to consider two cases: a robot has more DOFs than a Labanotation score and a Labanotation score has more DOFs than a robot. Original Labanotation can define three columns for an arm: upper arm, forearm, and hand instead of one column as shown in Fig. 1(a). However, due to the limitation of sensors, we may have to omit some of the columns. It often occurs to concatenate forearm and hand columns as an

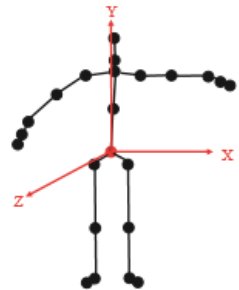

Fig. 5 A body coordinate system

Fig. 6 A simple robot to reproduce upper body motions

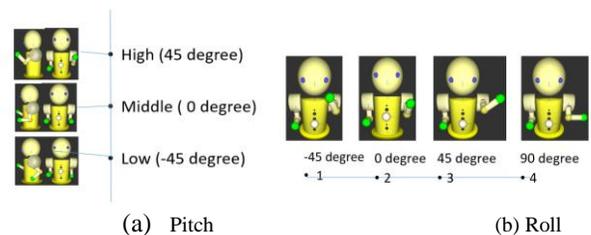

(a) Pitch          (b) Roll

Fig 6 All possible configuration of the left arm

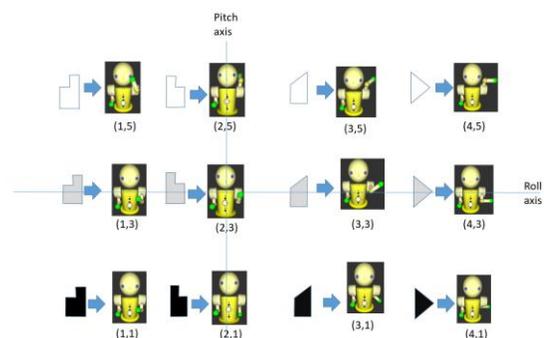

Fig. 7 Mapping Labanotation symbols to robot configurations

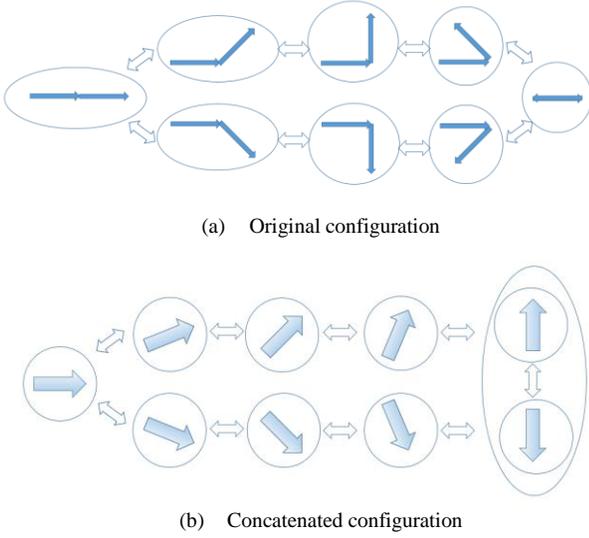

(a) Original configuration

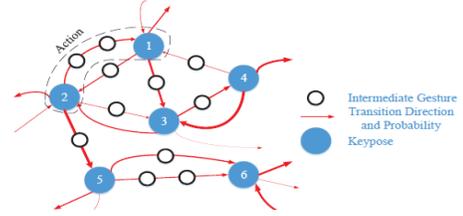

(b) Concatenated configuration

Fig 7 concatenation method

arm column as is the case for a Kinect sensor. When a robot has more DOFs than a Labanotation score due to the limitation of a sensor, we simply map the concatenate direction to two robot parts; a Labanotation symbol in the forearm column is map both to the robot's forearm and hand directions. For further complicated robots, we can apply the similar idea.

When a Labanotation score has more DOFs than a robot, we recursively combine adjacent Labanotation symbols into one symbol until the approximation is consistent to the robot DOFs. Since each body part are connected to each other and the Labanotation digitizes the direction in 45 degrees, possible configurations between two parts consist of eight cases: continue, foreword diagonal, orthogonal, backward diagonal, and reverse as shown in Fig. 7(a). Then, the reachable directions are 14 directions as shown in Fig 7(b). One singular case occurs at the reverse position. By considering the history of transition, either direction is selected. This eight cases occur to both directions and levels. This concatenation maps to robot parts directions. Fortunately, the Labanotation denotes each body parts separately. Only necessary depth of the recursive operation to be considered is three for upper body motion.

## B. Trajectory Generation

A task model only provides the start and end states represented by Labanotation symbols. For a robot movement, we need a trajectory to specify intermediate motions between two states. In this paper, we implemented an interpolation method. The intermediate trajectories are generated based on linear and cubic interpolation methods.

We can represent more complicated trajectory based on observation. The purpose of this paper is to propose task models based on Labanotation. However, we briefly explain dictionary construction based on observation. We plan to connect our observation module to a cloud computer and store those trajectories on the computer.

Given a pair of key poses, if intermediate motions are similar with existing ones, we just update the transition probability. Otherwise, we will add the intermediate motions as a new transition path and update the dictionary. By continuous observation, we can construct trajectories for a transition in a dictionary on a cloud computer.

Fig. 8 shows a constructed dictionary from motion analysis. Given a sequence of motions, each motion is represented by a pair of key poses. Based on the length of the motion and sampling intervals, possible intermediate gestures also are restored to guide the motion reconstruction.

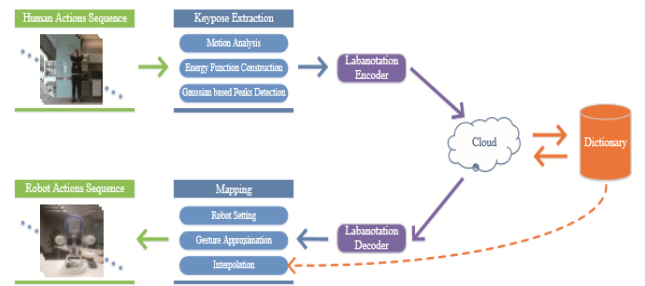

Fig 8 An example of a motion dictionary

## IV. EXPERIMENTS

### A. System Implementation

Fig.9 shows the overview of our system. In the human parts, key frames extraction and Labanotation encoder are implemented. A Kinect sensor is utilized for recording human movements. In this particular implementation, the cloud part is implemented on the same computer, which collects trajectories and constructs the dictionary. Fig 8(b) is a physical set up. Different from GR-001 robot [23], our own robot contains 9 DOFs, namely 1 to the body (yaw), 2 to the head (pitch and yaw), and 3 to each of two arms. Each DOF of the robot is driven by a servo motor (RS303MR). For robot control, command is received from computer via processing unit (RPU-19). After a certain latency, typically 5s, the similar movements are generated on the robots. However, once movements are learned, any number of performance can be done later.

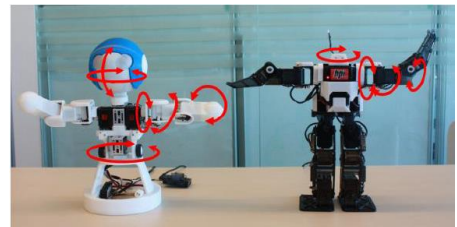

(a) System configuration

(b) Robots to be used

Fig. 9 Demonstration set up

## B. Evaluation

We will evaluate a couple of issues on demonstration. The first evaluation is how well the system demonstrates human motions. For that, we compare human motions with those robots. Given a human motion as the input, our robot system automatically extracts key frames and translates each movements into Labanotation symbols. Then, a task sequence is defined based on the learning-from-observation method. As for the output, the task sequence, along with the dictionary, is generated to guide robot motions. Mapping routines are designed by interpolating intermediate motions between each pair of key frames. Here, the dictionary is first used to select partial intermediate motions. Then, other intermediate motions are generated by using the linear interpolation method.

We first compare human motions with robot motions in one pair of key frames, as shown in Fig. 10(a). The top and bottom rows depict original human and robot motions, respectively. The gestures surrounded by red dotted boxes are those at key frames. As is analyzed above, Labanotation actually is the representation method that divides the motion space into discrete directions and levels. Thus, gestures with slight differences can share the same Labanotation, as shown in the figures of the forth and the fifth columns. In addition, since robots' motion speeds are different, captured intermediate gestures between the two key poses are slightly different. To validate the hardware independency of our robot system, we show more key poses and less intermediate gestures, and analyze difference between different robot platforms. As shown in Fig. 10(b), for key poses, the two robots can effectively mimic human gestures, though visual results are slightly different due to their different DOFs. Observing intermediate gestures, we find that different DOFs also generate different intermediate gestures, even though we apply the same interpolation method. Considering that mapping routines of the two robots are different, difference in robot motions just demonstrates the hardware independency of our robot system, especially upper body task models.

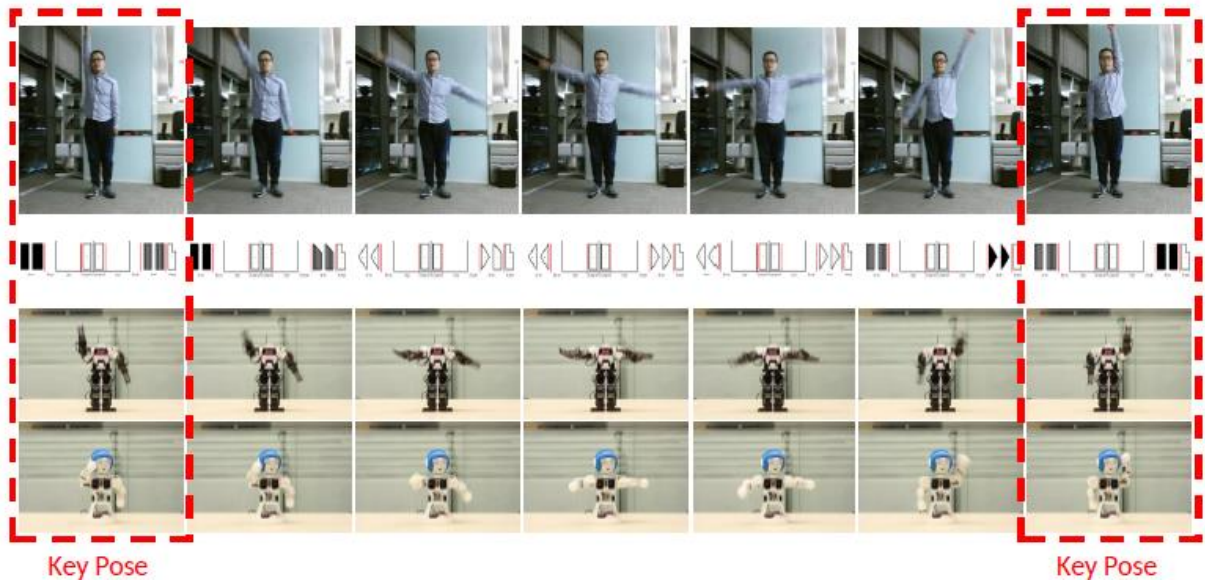

(a) Intermediate gesture evaluation

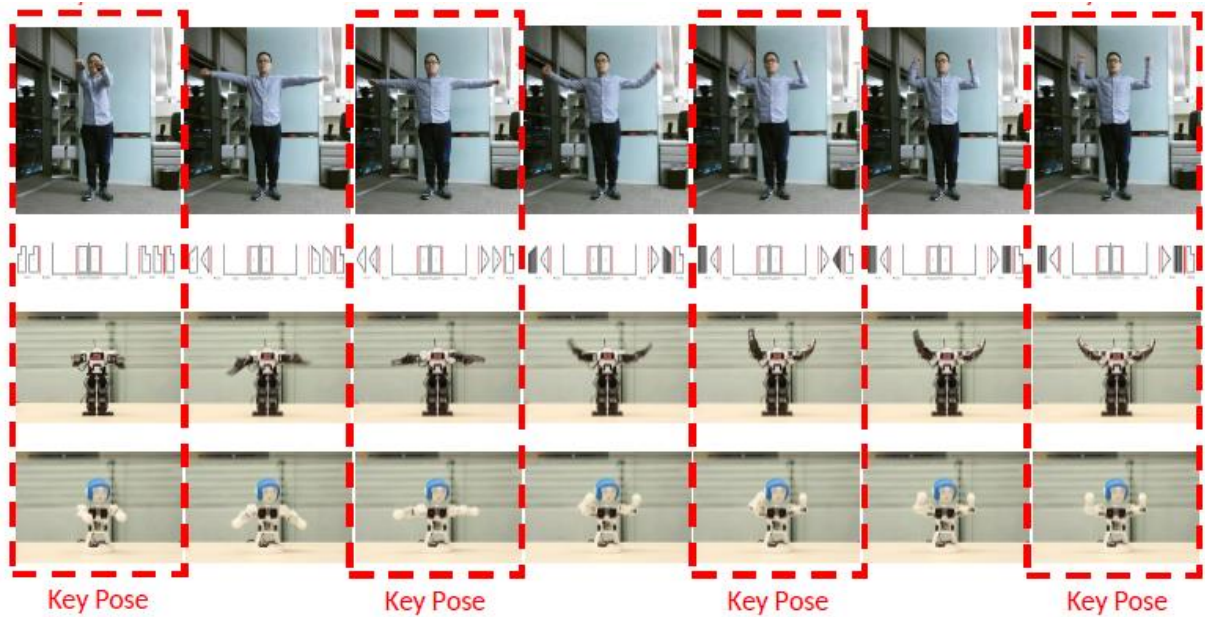

(b) Pairwise comparison

Fig. 10 Demonstration results

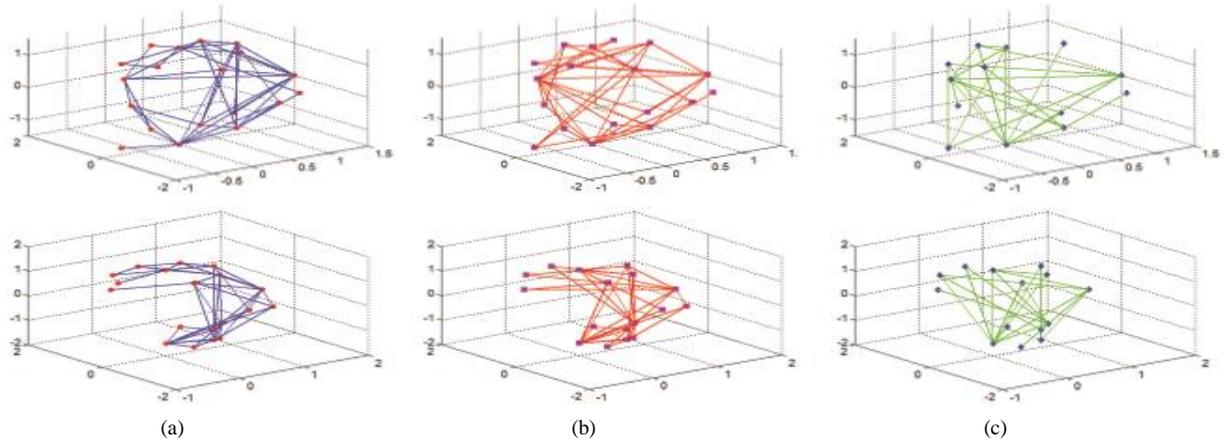

Fig 11. Comparision of motion trajectories of the left wrist joint (first row) and the right wrist joint (second row): (a) motion trajectories according to all gestures in a motion sequence, (b) motion trajectories based on key poses and intermediate gestures (directly selected from the constructed dictionary), and (c) motion trajectories based on the key poses and linear interpolation method.

To show the performance of different interpolation methods on routine mapping, we construct motion trajectories according to different methods, and compare these trajectories with original human motion trajectories as shown in Fig. 11. To construct motion trajectories, we first draw XYZ coordinates of the joint (for both key poses and intermediate gestures from the dictionary) as red points. Then, we link two points by a solid line if the two points represents two successive gestures in human or robot motions. It needs to be pointed out that constructed motion trajectories are not real motion trajectories, since one single point might represent a couple of repeat gestures. Although constructed motion trajectories in Fig. 10 (a) are relatively different from real human motion trajectories, we still can use such these trajectories to compare the performance of different interpolation methods. Obviously, by taking partial intermediate gestures in the dictionary to help guide the interpolation, robot can effectively approximate original human motions. For the linear interpolation method, intermediate gestures are generated based on keyposes, so that robot motion probably would be quite different from human motions.

## V. Conclusion

This paper proposes the Labanotation as the basis for defining task models of learning-from-observation in the domain of upper body motion. We construct a robot system to observe and mimic human performance, especially upper body motions. By observing human movements, we first extract key frames, where one part of a human body briefly stops, via the analysis of upper-body motions. To accomplish the hardware independency, we introduce Labanotation as the basic representations for the task models. Since Labanotation is independent from robot hardware, task models, only related to key poses, can be executed on difference robot platforms by employing different mapping routines. We implemented the proposed systems on our own robot and the GR001 robots.

In this paper, we do not focus on skill modeling, namely trajectory generation. We simply interpolate the intermediate gestures. Depending on the performer, the trajectories are slightly different each other. How to describe such trajectory difference is an open issue. Laban also proposed the Laban effort graph to represent how to generate trajectories as well as speed along a trajectory by symbolic representations as Sudden, Smooth, Direct, and Indirect. In future, we will implement this trajectory specification based on observation and characterization based on Laban efforts for cloud robots.